\documentclass{bmvc2k}

\usepackage{amsmath,amssymb} 
\usepackage{color}
\usepackage{adjustbox}
\usepackage{bm}
\usepackage{algorithm}
\usepackage{algpseudocode}
\usepackage{wrapfig}

\title{Domain Adaptation for the Segmentation of Confidential Medical Images}

\addauthor{Serban Stan}{ }{1,2}
\addauthor{Mohammad Rostami}{ }{1,2}

\addinstitution{
Information Sciences Institute\\
 University of Southern California\\
 Los Angeles, USA
}
\addinstitution{
 Department of Computer Science\\
 University of Southern California\\
 Los Angeles, USA
}

\runninghead{Stan and Rostami}{Adaptation for Medical Segmentation}

\begin{document}

\maketitle

\begin{abstract}
Convolutional neural networks (CNNs) have led to significant improvements in the semantic segmentation of images. When source and target datasets come from different modalities, CNN performance suffers due to domain shift. In such cases data annotation in the target domain becomes necessary to maintain model performance. To circumvent the re-annotation process, unsupervised domain adaptation (UDA) is proposed to adapt a model to new modalities using solely unlabeled target data. Common UDA algorithms require access to source domain data during adaptation, which may not be feasible in medical imaging due to data sharing restrictions. In this work, we develop an algorithm for UDA where the source domain data is inaccessible during target adaptation. Our approach is based on encoding the source domain information into an internal distribution that is used to guide adaptation in the absence of source samples.  We demonstrate the effectiveness of our algorithm by comparing it to state-of-the-art medical image semantic segmentation approaches on two medical image semantic segmentation datasets. 
\end{abstract}

\section{Introduction}
\label{sec:introduction}

 Employing CNNs in semantic segmentation tasks has been proven to be extremely helpful in various applications, including object tracking~\cite{yilmaz2006survey,bertinetto2016fully,Zhu_2018_ECCV}, self-driving cars~\cite{Kim_2017_ICCV,Hecker_2018_ECCV}, and medical image analysis~\cite{ker8241753deep,shen2017deep,AYACHE2017xxiii,KAZEMINIA2020101938}. This success, however, is conditioned on the availability of huge manually annotated datasets to supervise the training of state-of-the-art (SOTA)  network structures~\cite{rostami2018crowdsourcing,Toldo_Maracani_Michieli_Zanuttigh_2020}. This condition is not always realized in practice, especially in fields such as medical image segmentation, where annotating data requires the input of trained experts and privacy regulations make sharing data for crowd-sourcing extremely restricted, and at times impossible. A characteristic of data in the area of medical image segmentation is the existence of \textit{domain shift} between  different imaging modalities, which stems from using imaging  devices based on totally different electromagnetic principles, e.g., CT vs MRI.    When domain gap exists between the distributions of the training (source) and the testing (target) data, the performance of CNNs can degrade significantly. This  makes continual data annotation necessary for maintaining model performance.
 
 Domain shift is a major area of concern, as data annotation is a challenging procedure even for the simplest semantic segmentation tasks~\cite{estimating2011liu}. Annotating medical images is also expensive, as annotation can be performed only by physicians, who undergo years of training to obtain domain expertise.    
Unsupervised domain adaptation (UDA) is a learning setting aimed at reducing \textit{domain gap} without data annotation in the target domain. The goal is to adapt a source-trained model for improved generalization in the target domain using solely unannotated data~\cite{ghifary2016deep,venkateswara2017deep,saito2018maximum,zhang2018collaborative}. The core idea in UDA is to achieve knowledge transfer from the source domain to the target domain by aligning the  latent features of the two domains in an embedding space. This idea has been implemented either using adversarial learning~\cite{hoffman2018cycada,dou2019pnp,tzeng2017adversarial,bousmalis2017unsupervised}, directly minimizing the distance of distributions of the latent features with respect to a probability metric~\cite{chen2019progressive,sun2017correlation,lee2019sliced,rost2021unsupervised},  or a combination of the two~\cite{choi2019self,sankaranarayanan2018generate}. 

While existing UDA algorithms have been successful in reducing cross-domain gap, the vast majority of these approaches require sharing data between  the source and target domains to enforce distribution alignment. This requirement limits the applicability of most existing works when sharing data may not be possible, e.g., sharing data is heavily regulated in healthcare domains due to the confidentiality of patient data and from security concerns. Until recently, there has been little exploration of UDA when access to the source domain is limited~\cite{kundu2020universal,saltori2020sf,qiu2021sourcefree,yang2021generalized,rostami2021domain}. These recent works benefit from generative adversarial learning to maintain source distribution information. However, addressing UDA for classification tasks limits the applicability of such methods to the problem of organ semantic segmentation~\cite{zhao2019knowledge}. A similar problem is encountered with UDA for street semantic segmentation \cite{Kundu_2021_ICCV}, given medical devices produce data distributions requiring additional preparation with large background areas \cite{zhou2019prioraware}. Recent medical works propose adaptation without source access via entropy minimization ~\cite{bateson2022source,BatesonSFDA}, but these methods are susceptible to degenerate solutions.

\textbf{Contribution:}  we develop a UDA algorithm for the semantic segmentation of medical images when sharing data is infeasible due to confidentiality or security concerns. Our approach is able to reduce domain gap without having direct access to the source data during adaptation. We learn the internal distribution for the source domain, and transfer knowledge between the source and target domains by distribution alignment between the learned internal distribution and the latent distribution of features of the target domain. We validate our algorithm on two medical image segmentation datasets, and observe comparable performance to SOTA methods based on joint training.

\section{Related Work}
\label{sec:relatedwork}

SOTA semantic segmentation algorithms  use deep neural network architectures to exploit large annotated datasets \cite{long2015fully,noh2015learning,lin2017refinenet,lecun2015deep}. These approaches are based on training a CNN encoder using manually annotated segmentation maps to learn a latent embedding of the data. An up-sampling decoder combined with a classifier is then used to infer pixel-wise estimations for the true semantic labels. Performance of such methods is  high when  large amounts of annotated data are available for supervised training. However, these methods are not suitable when the goal is to transfer knowledge between different domains~\cite{saito2018maximum,pan2019transferrable}. 
Model adaptation from a fully annotated source domains to a target domains has been explored in both semi-supervised and unsupervised settings. Semi-supervised approaches rely on the presence of a small number of annotated target data samples~\cite{motiian2017few,zhang2019few}. For example, a weakly supervised signal on the target domain can be obtained using bounding boxes. However, manual data annotation of a small number of images is still a considerable bottleneck in the area of medical imaging because only trained professionals can perform this task. For this reason, UDA algorithms are more appealing for healthcare applications.

UDA approaches have explored two main strategies to reduce the domain gap. A large number of works rely on generative adversarial networks (GANs)~\cite{Zhang_2018_CVPR,10.1007/978-3-319-59050-9_47}. The core idea is to use a GAN loss such that   data points from both domains can be mapped into a domain-invariant embedding space~\cite{hoffman2018cycada}.
To this end, a cross-domain  discriminator network is trained to classify whether data embeddings correspond to the source or target domain. An encoder network attempts to fool the discriminator by producing domain agnostic representations for source and target data points. Following this alternating optimization process, a classifier trained using source domain encodings produced by the encoder network would also generalize in the target domain~\cite{Ma_2019_CVPR,dou2018unsupervised}. The weakness of GANs is mode collapse, which requires careful fine-tuning and selection of hyper-parameters in order to be overcome.

Other UDA approaches aim to directly align the distributions of the two domains in a shared embedding space~\cite{lee2019sliced,rostami2021transfer,stan2021unsupervised}. A shared encoder network is used to generate latent features for both domains. A common latent feature space is achieved by minimizing a suitable probability distance metric between the source and target embeddings \cite{drossos2019unsupervised,LE2019249,rostami2019deep,liu2020pdam}. Selecting proper distance metrics has been the major focus of research for these approaches. Optimal transport has been found particularly suitable for deep learning based UDA ~\cite{courty2016optimal}. We utilize the Sliced Wasserstein Distance (SWD)~\cite{lee2019sliced,rostami2021lifelong} variant of the optimal transport with similar properties, but which allows for fast gradient based optimization.

The above mentioned sets of approaches have been found helpful in various medical semantic segmentation applications~\cite{huo2018adversarial,zhang2018task,chen2018semantic,kamnitsas2017unsupervised}. However, both strategies require direct access to source domain data for computing the loss functions. To relax this requirement, UDA has been recently explored in a source-free setting to address scenarios where the source domain is not directly accessible~\cite{kundu2020universal,saltori2020sf}.
Both Kundu et al.~\cite{kundu2020universal} and Saltori et al.~\cite{saltori2020sf} target image classification, and benefit from generative adversarial learning to generate  pseudo-data points that are similar to the source domain data in the absence of actual source samples. While both approaches are suitable for classification problems,  extending them to semantic segmentation of medical images is not trivial. First, training models that can generate realistic medical images is considerably more challenging due to importance of fine details. Second, one may   argue that if generated images are too similar to real images, the information confidentiality of patients in the training data may still be compromised. Our work is based on using  a dramatically different approach. We develop a source-free UDA algorithm that performs the distribution alignment of two domains in an embedding space by using an intermediate internal distribution to relax the need for source   data. 

\section{Problem Formulation}
\label{sec:problemformulation}

\begin{figure*}[!htb]
    \centering
     \includegraphics[width=.8\textwidth]{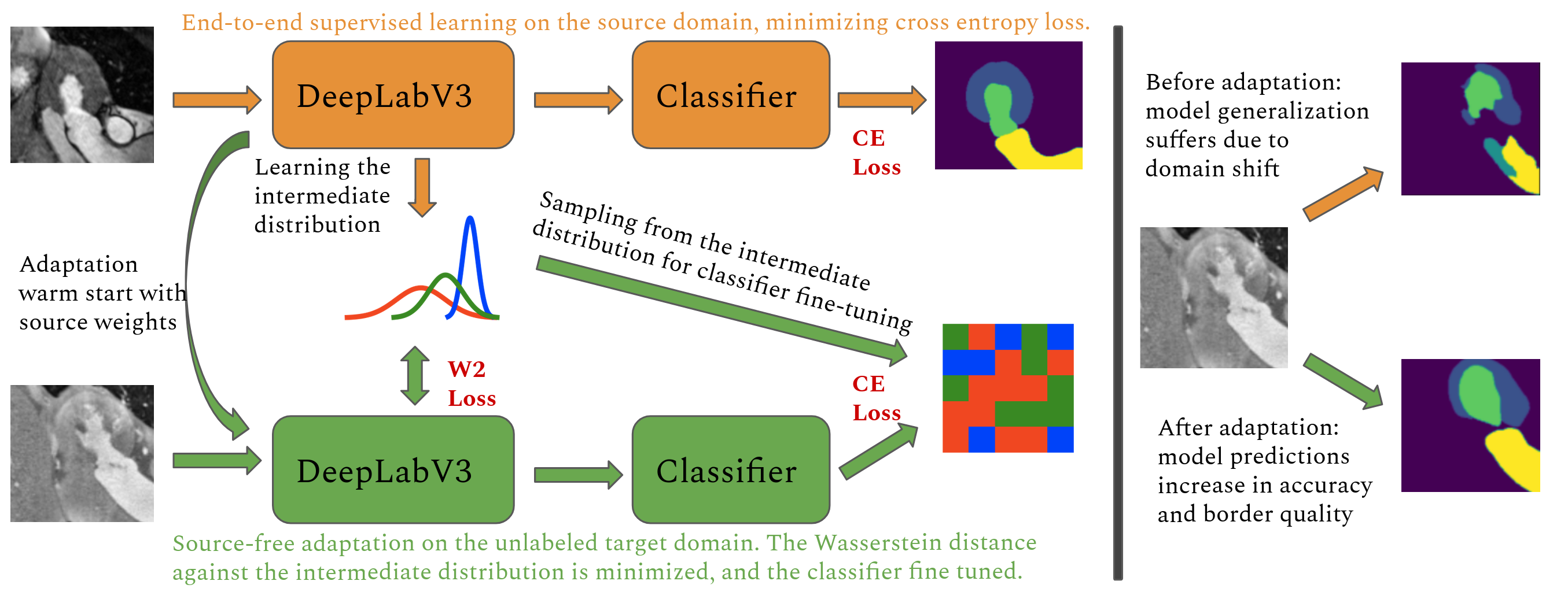}
    \caption{\small Proposed method: We first perform supervised training on source MR images. Using the source embeddings we characterize an internal distribution via a GMM distribution in the latent space. We then perform source free adaptation by matching the embedding of the target CT images to the learnt GMM distribution, and fine tuning of the classifier on GMM samples. Finally, we verify the improved performance that our model gains from model adaptation. }
     \label{figure:problem-formulation}
     \vspace{-5mm}
\end{figure*}

Consider a source domain $D^S=(X^S,Y^S)$ with annotated data and a target domain $D^T=(X^T)$ with unannotated data that despite having different input spaces $X^S$ and $X^T$, e.g., due to using different medical imaging techniques, share the same segmentation map space $Y$, e.g., the same tissue/organ classes.  Following the standard UDA pipeline, the goal is to learn a segmentation mapping function for the target domain by transferring knowledge from the source domain. To this end, we must learn a function $f_{\theta}(\cdot):\{X^S\cup X^T\} \rightarrow \{Y\}$ with learnable parameters $\theta$, e.g., a deep neural network, such that given an input image $x^*$, the function returns a  segmentation mask $\hat y$ that best approximates the ground truth segmentation mask $y^*$. Given the annotated training dataset  $\{(x^s,y^s)\}_{i=1}^N$ in the source domain, it is straightforward to train a segmentation model that generalizes well in the source domain   through solving an empirical risk minimization (ERM) problem, i.e., $ \hat\theta = \arg \min_\theta \frac{1}{N} \sum_{i=1}^N \mathcal{L} (y^s, f_\theta(x^s)))$, where $\mathcal{L}$ is a proper loss function, e.g., the pixel-wise cross-entropy loss, defined as $\mathcal{L}_{ce} (y^*, \hat y) = - \sum_{i=1}^W \sum_{j=1}^H \sum_{k=1}^K y^*_{ijk} \log \hat y_{ijk}$. Here, $K$ denotes the number of segmentation classes, and $W,H$ represent the  width and the height the input images, respectively. Each pixel label $y^*_{ij}$ will be represented as a one hot vector of size $K$  and $\hat y_{ij}$ is the prediction vector which assigns a probability weight to each label. Due to the existence of domain gap across the two domains, i.e. discrepancy between the source domain distribution $p^s(X)$ and the target domain distribution $p^t(X)$, the source-trained model using ERM may generalize poorly in the target domain. We want to benefit from the information encoded in the target domain unannotated dataset  $\{x^t\}_{i=1}^M$ to improve the model generalization in the target domain further.

We follow the common strategy of domain alignment in a shared embedding space to adress UDA. Consider our model $f$ to be a deep convolutional neural network (CNN). Let $f = \phi \circ \chi \circ \psi$, where $\psi(\cdot):\mathbb{R}^{W\times H \times C} \rightarrow \mathbb{R}^{U \times V}$ is a CNN encoder, $\chi(\cdot):\mathbb{R}^{U\times V} \rightarrow \mathbb{R}^{W \times H \times K}$ is an up-scaling CNN decoder, and $\phi(\cdot) : \mathbb{R}^{W \times H \times K} \rightarrow \mathbb{R}^{W \times H  \times K}$ is a classification network that takes as inputs latent space representations and assigns label-probability values. We model the shared embedding space as the output space of the sub-network  $\chi\circ\psi(\cdot)$.
Solving UDA reduces to ensuring the source and target embedding distributions are aligned in the embedding space. This translates into minimizing the distributional discrepancy between the $\chi\circ\psi(p^s(\cdot))$ and $\chi\circ\psi(p^t(\cdot))$ distributions. 
A large group of UDA algorithms \cite{wu2018_dcan_eccv,zhang2017_curriculum} select a  probability distribution metric $D(\cdot,\cdot)$, e.g. SWD or KL-divergence, and then use the source and the target domain data points, $X^S=[x^s_1,\ldots, x^s_N]$ and $X^T=[x^t_1,\ldots,x^t_N]$, to minimize the loss term  $D(\chi\circ\psi(p^s(\cdot)),\chi\circ\psi(p^t(\cdot)))$ as a regularizer. However, this will   constrain the user to have access to the source domain data to compute $D(\chi\circ\psi(p^s(\cdot)),\chi\circ\psi(p^t(\cdot)))$ that couples the two domains. 
We provide a solution to align the two   domains without sharing the source domain data, that benefits from an intermediate probability distribution.

\section{Proposed Algorithm}
\label{sec:proposedalgorithm}

Our proposed approach is based on using the internal distribution $\mathcal{P_Z}$ as a surrogate for the learned distribution of the source domain in the embedding space.  Upon training $f_\theta$ using ERM,  the embedding space would become discriminative for the source  domain. This means that the source distribution in the embedding space will be a multimodal distribution, where each mode denotes one of the classes. This distribution can be modeled as a  Gaussian Mixture Model (GMM). To develop a source-free UDA algorithm, we can draw random samples from the GMM and instead of relying on the source data, align the target domain distribution with the internal distribution in the embedding space. In other words, we estimate the term $D(\chi\circ\psi(p^s(\cdot)),\chi\circ\psi(p^t(\cdot)))$ with $D(\mathcal{P_Z}(\cdot),\chi\circ\psi(p^t(\cdot)))$ which does not depend on source samples. We use SWD as the distribution metric for minimizing the domain discrepancy. A visual concept-level description for  our approach is presented in Figure \ref{figure:problem-formulation}.

\textbf{The intermediate distribution.} The function $\psi \circ \chi$  transforms the input distribution $p^s(\cdot)$ to the internal distribution $\mathcal{P_Z}(\cdot) = \chi\circ\psi(p^s(\cdot))$ based on which the classifier $\phi$ assigns labels. This distribution will have $K$ modes. Our   idea is to approximate $\mathcal{P_Z}(\cdot)$ via a GMM with $\omega \times K$ components, with $\omega$ components for each of the $K$ semantic classes:
\begin{equation*}
    \small
    \mathcal{P_Z}(z) = \sum_{c=1}^{\omega K} \alpha_c p_c(z)
        = \sum_{c=1}^{\omega K} \alpha_c \mathcal{N} (z | \mu_c, \Sigma_c),
\end{equation*}
where $\alpha_c$ represents the mixture probabilities, $\mu_c$ represents the mean of the Gaussian $c$, and $\Sigma_c$ is the covariance matrix of the $c^{th}$ component. Under the above representation, each semantic class $k \in \{1 \dots K\}$ will be represented by $\omega$ components: $(k-1) \omega + 1 \dots k \omega$.
When the network $f$ is trained on the source domain, we can estimate the GMM parameters class-conditionally from the latent features obtained from the source training samples $\{ \phi(\chi(x^s))_{ijt}, y^s_{ij} \}$. Once class specific latent embeddings are computed via access to the labels $Y^S$, we estimate the corresponding $\omega$ components using the EM algorithm.  

\textbf{Sample selection.} To improve class separations in the internal distribution $\mathcal{P}_Z$, we only use high-confidence samples in each class for estimating parameters of $p_c(\cdot)$. We use a confidence threshold parameter $\rho$, and discard all samples for which the classifier confidence on its prediction $p_{ij}$ is strictly less than $\rho$. This step helps cancel out class outliers. 
Let $\mathcal{S}_\rho = \{(x_{ij}^s, y_{ij}^s) | \max\phi(\chi(\psi(x_{ij}))) > \rho\}$ be the source data pixels on which the classifier $\phi$ assigns confidence greater than $\rho$. Also, let $\mathcal{S}_{\rho,k} = \{ (x,y) | (x,y) \in \mathcal{S}_\rho, y = k\}$. Then, for each class $k$ we generate empirical estimates for the $\omega$ components defined by triplets $(\hat \alpha_{\omega (k - 1) + 1}, \hat \mu_{\omega (k - 1) + 1}, \hat \Sigma_{\omega (k - 1) + 1} ) \dots (\hat \alpha_{\omega k}, \hat \mu_{\omega k}, \hat \Sigma_{\omega k})$ by applying EM to $\mathcal{S}_{\rho,k}$ data points. 



\textbf{The adaptation loss.} Given the estimated internal distribution parameters $\hat\alpha, \hat\mu, \hat\Sigma$, we can perform domain alignment. Adapting the model should lead to the target latent distribution $ \chi(\psi(p^t(X)))$ matching the distribution $\mathcal{P}_Z$ in the embedding space. 
To this end, we can generate a pseudo-dataset $D^P = (Z^P,Y^P)$ by drawing samples from the GMM and aligning $\chi(\psi(X^T))$ with $Z^P$ to reduce the domain gap. The alignment loss can then be formalized as:

\begin{equation}
\small
\begin{split}
    \mathcal{L}_{adapt} = \mathcal{L}_{ce} (\phi(Z^P), Y^P) + \lambda \mathcal{D} (\chi(\psi(X^T)), Z^P)
\end{split}
\label{eq:AdaptLoss}
\end{equation}  
The first term in Eq.~\ref{eq:AdaptLoss} involves fine-tuning the classifier on samples from the pseudo-dataset $(Z^P,Y^P)$ to ensure that it would continue to generalize well. The second term enforces the distributional alignment.  Since the source samples are not used in Eq.~\ref{eq:AdaptLoss},   data confidentiality   will also be preserved. 
The last ingredient for our approach is selection of the distance metric $\mathcal{D}(\cdot,\cdot)$. We used SWD for this purpose. The pseudocode for our approach, called Source Free semantic Segmentation (SFS), is presented in Alg. ~\ref{SSUDAalgorithmSS}.

 \begin{wrapfigure}{R}{0.5\textwidth}
  \begin{minipage}{0.5\textwidth}
 \vspace{-25mm}
\begin{algorithm}[H]
\caption{$\mathrm{SFS}\left (\lambda , \rho, \omega \right)$\label{SSUDAalgorithmSS}} 
 {\small
    \begin{algorithmic}[1]
    \State \textbf{Initial Training}: 
    \State \hspace{2mm}\textbf{Input:} Source   dataset $\mathcal{D}^{\mathcal{S}}=(\bm{X}^{\mathcal{S}},  \bm{Y}^{\mathcal{S}})$,
    \State \hspace{4mm}\textbf{Training on Source Domain:}
    \State \hspace{4mm} $\hat{ \theta}_0=\arg\min_{\theta}\sum_i \mathcal{L}(f_{\theta}(\bm{x}_i^s),\bm{y}_i^s)$
    \State \hspace{2mm}  \textbf{Internal Distribution Estimation:}
    \State \hspace{4mm} Set $\rho=.97$, compute $\mathcal{S}_{\rho, k}$, and   estimate $\hat\alpha_j, \hat\mu_j,$ and $\hat\Sigma_j$ class conditionally via EM
    \State \textbf{Model Adaptation}: 
    \State \hspace{2mm} \textbf{Input:} Target dataset $\mathcal{D}^{\mathcal{T}}=(\bm{X}^{\mathcal{T}})$
    \State \hspace{2mm} \textbf{Pseudo-Dataset Generation:} 
    \State   $\mathcal{D}^{\mathcal{P}}=(\textbf{Z}^{\mathcal{P}},\textbf{Y}^{\mathcal{P}})=([\bm{z}_1^p,\ldots,\bm{z}_N^p],[\bm{y}_1^p,\ldots,\bm{y}_N^p])$, where: $\bm{z}_i^p\sim \mathcal{P}_Z(\bm{z}), 1\le i\le N_p$
    \State \hspace{2mm} \textbf{for} $itr = 1 \rightarrow ITR$ \textbf{do}
    \State \hspace{4mm} Draw batches from $\mathcal{D}^{\mathcal{T}}$ and $\mathcal{D}^{\mathcal{P}}$
    \State \hspace{4mm} Update the model by solving Eq.~\eqref{eq:AdaptLoss}
    \State \hspace{2mm} \textbf{end for}
    \end{algorithmic}}
\end{algorithm} 
 \end{minipage}
 \vspace{-2mm}
  \end{wrapfigure} 

\section{Experimental Validation}
\label{sec:experimentalvalidation}

\subsection{Datasets}


\textbf{Multi-Modality Whole Heart Segmentation Dataset (MMWHS)~\cite{ZHUANG201677}:}  this dataset consists of multi-modality whole heart images obtained on different imaging devices at different imaging sites. Segmentation maps are provided for $20$ MRI 3D heart images and $20$ CT 3D heart images which have domain gap. Following the UDA setup, we use the MRI images as the source domain and CT images as the target domain. We perform UDA with respect to four of the available segmentation classes: ascending aorta (AA), left ventricle blood cavity (LVC), left atrium blood cavity (LAC), myocardium of the left ventricle (MYO). 

We will use the same experimental setup and parsed dataset used by Dou et al.~\cite{dou2018unsupervised} for fair comparison. For the MRI source domain we use augmented samples from $16$ MRI 3D instances. The target domain consists of augmented samples from of $14$ 3D CT images, and we report results on $4$ CT instances, as proposed by Chen et al.~\cite{chen2020unsupervised}. Each 3D segmentation map used for assessing test performance is normalized to have zero mean and unit variance. 

\textbf{CHAOS MR $\rightarrow$ Multi-Atlas Labeling Beyond the Cranial Vault:} the second domain adaptation task consists of data frames from two different dataset. As source domain we, consider the the 2019 CHAOS MR dataset \cite{CHAOSdata2019}, previously used in the 2019 CHAOS Grad Challenge. The dataset consists of both MR and CT scans with segmentation maps for the following abdominal organs: liver, right kidney, left kidney and spleen. Similar to \cite{chen2020unsupervised} we use the T2-SPIR MR images as the source domain. Each scan is centered to zero mean and unit variance, and values more than three standard deviations away from the mean are clipped. In total, we obtain $20$ MR scans, $16$ of which we use for training and $4$ for validation.
The target domain is represented by the dataset which was presented in the Multi-Atlas Labeling Beyond the Cranial Vault MICCAI 2015 Challenge \cite{landman2015multiatlas}. We utilize the $30$ CT scans in the training set which are provided segmentation maps, and use $24$ for adaptation and $6$ for evaluating generalization performance. The value range in the CT scans was first clipped to $[-125,275]$ HU following literature \cite{zhou2019prioraware}. The images were re-sampled to an axial view size of $256\times256$. Background was then cropped such that the distance between any labeled pixel and the image borders is at least $30$ pixels, and scans were again resized to $256\times256$. Finally, each 3D scan was normalized independently to zero mean and unit variance, and values more than three standard deviation from the mean were clipped. Data augmentation was performed   on both the training MR and training CT instances using: (1) random rotations  of up to $20$ degrees, (2) negating pixel values, (3) adding random Gaussian noise, (4) random cropping.

Both of the above problems involve $3D$ scans. However our network encoder architecture receives  $2D$ images at its input, where each image consists of three channels. To circumvent this discrepancy, we follow the frame-by-frame processing methodology by Chen et al.~\cite{chen2019synergistic}. We convert higher dimensional features into $2D$ images by creating images from groups of three consecutive scan slices, and using them as labels for the segmentation map of the middle slice.
Implementation details re included in the Appendix.

\vspace{-3mm}

\subsection{Evaluation Methodology}

Following the medical image segmentation literature, we use two main metrics for evaluation:  dice coefficient and average symmetric surface distance (ASSD). The Dice coefficient is a popular choice in medical image analysis works and measures semantic segmentation quality~\cite{chen2019synergistic,chen2020unsupervised,zhou2019prioraware}. It is used for direct evaluation of segmentation accuracy. The ASSD is a metric which has been  used~\cite{SUN201758,chen2020unsupervised,DOU201740} to assess the quality of borders of predicted segmentation maps which are important for diagnosis. A good segmentation will have a large Dice coefficient and low ASSD value, the desirability of a result being application dependant.

We compare our approach to other state-of-the-art techniques developed for unsupervised medical image segmentation. We compare against adversarial approaches PnP-AdaNet~\cite{dou2019pnp}, SynSeg-Net~\cite{Huo_2019}, AdaOutput~\cite{Tsai_adaptseg_2018}, CycleGAN~\cite{zhu2020unpaired}, CyCADA~\cite{hoffman2018cycada}, SIFA~\cite{chen2020unsupervised}, ARL-GAN~\cite{chen2020anatomy}, DSFN~\cite{ijcai2020-455}, SASAN~\cite{tomar2021self}, DSAN~\cite{han2021deep}. These works are recent methods for semantic segmentation that serve as \textbf{upper bounds} for our approach, as we do not process the source domain data directly. We reiterate the advantage of our method is to preserve the confidentiality of patient data, and we do not claim best performance. We also compare against GenAdapt \cite{Kundu_2021_ICCV}, a SOTA street semantic segmentation method that is not tuned for the medical field. Finally, we also evaluate our model against AdaEnt~\cite{BatesonSFDA} and AdaMI~\cite{bateson2022source}, two recent source-free approaches designed for medical semantic segmentation, and observe our methods outperforms both these techniques.
Our code is available as a supplementary material.

\vspace{-4mm}

\subsection{Quantitative and  Qualitative Results}

Tables \ref{table:results-cardiac} and \ref{table:results-abdomen} summarize the segmentation performance for our method along with other baselines. As mentioned, when compared to other UDA approaches our method has the additional benefit of not violating data confidentiality on the source and target. This means most other approaches should serve as \textbf{upper bounds} for our algorithm, as they do not enforce restrictions for jointly accessing source and target data. We also compare against a recent street semantic segmentation algorithm \cite{Kundu_2021_ICCV} to verify whether real world adaptation approaches are at a disadvantage due to the specificity of medical data. We observe this approach has indeed lowest performance out of the considered methods. 
On the MMWHS dataset we achieve SOTA performance on class AA. We obtain the highest Dice score out of the considered methods, due to our high average performance on all classes. The ASSD score is competitive with other approaches, the best such score being observed for GAN based methods. This shows our domain alignment approach successfully maps each class in the target embedding to its corresponding vicinity using the internal distribution. For the abdominal task we observe similar trends. We achieve SOTA performance on class \textit{Liver}, and competitive performance on the other classes. These results suggest that our method offers the possibility of domain adaptation with competitive performance.

\begin{table}[ht!]
  \begin{adjustbox}{center}
    \scalebox{.77}{
    \small
    \setlength\tabcolsep{1.5pt} 
        \begin{tabular} { |c|cccc|c|cccc|c| }
            \hline
            &\multicolumn{5}{c|}{Dice} &\multicolumn{5}{c|}{Average Symmetric Surface Distance} \\ \hline
            Method &AA &LAC &LVC &MYO &Average &AA &LAC &LVC &MYO &Average \\ \hline
            Source-Only &28.4 &27.7 &4.0 &8.7 &17.2 &20.6 &16.2 &N/A &48.4 &N/A \\ \hline
            Supervised$^*$ &88.7 &89.3 &89.0 &88.7 &87.2 &2.6 &4.9 &2.2 &1.6 &3.6 \\ \hline
            GenAdapt$^*$ \cite{Kundu_2021_ICCV} &57 &51 &36 &31 &43.8 &N/A &N/A &N/A &N/A &N/A \\ \hline
            PnP-AdaNet \cite{dou2019pnp} &74.0 &68.9 &61.9 &50.8 &63.9 &12.8 &6.3 &17.4 &14.7 &12.8 \\ \hline
            SynSeg-Net \cite{Huo_2019} &71.6 &69.0 &51.6 &40.8 &58.2 &11.7 &7.8 &7.0 &9.2 &8.9 \\ \hline
            AdaOutput \cite{Tsai_adaptseg_2018} &65.2 &76.6 &54.4 &43.3 &59.9 &17.9 &5.5 &5.9 &8.9 &9.6 \\ \hline
            CycleGAN \cite{zhu2020unpaired} &73.8 &75.7 &52.3 &28.7 &57.6 &11.5 &13.6 &9.2 &8.8 &10.8 \\ \hline
            CyCADA \cite{hoffman2018cycada} &72.9 &77.0 &62.4 &45.3 &64.4 &9.6 &8.0 &9.6 &10.5 &9.4 \\ \hline
            SIFA \cite{chen2020unsupervised} &81.3 &79.5 &73.8 &61.6 &74.1 &7.9 &6.2 &5.5 &8.5 &7.0 \\ \hline
            ARL-GAN \cite{chen2020anatomy} &71.3 &80.6 &69.5 &\textbf{81.6} &75.7 &6.3 &5.9 &6.7 &6.5 &6.4 \\ \hline
            DSFN \cite{ijcai2020-455} &84.7 &76.9 &79.1 &62.4 &75.8 &N/A &N/A &N/A &N/A &N/A \\ \hline
            SASAN \cite{tomar2021self} &82.0 &76.0 &82.0 &72.0 &78.0 &\textbf{4.1} &8.3 &\textbf{3.5} &\textbf{3.3} &\textbf{4.9} \\ \hline
            DSAN \cite{han2021deep} &79.9 &\textbf{84.7} &\textbf{82.7} &66.5 &78.5 &7.7 &6.7 &3.8 &5.6 &5.9 \\ \hline
            AdaEnt$^*$ \cite{BatesonSFDA} &75.5 &71.2 &59.4 &56.4 &65.6 &8.5 &7.1 &8.4 &8.6 &8.2 \\ \hline
            AdaMI$^*$ \cite{bateson2022source} &83.1 &78.2 &74.5 &66.8 &75.7 &5.6 &\textbf{4.2} &5.7 &6.9 &5.6 \\ \hline
            \hline 
            \textbf{SFS$^*$} &\textbf{88.0} &83.7 &81.0 &72.5 &\textbf{81.3} &6.3 &7.2 &4.7 &6.1 &6.1 \\
            \hline
          \end{tabular}
    }
  \end{adjustbox}
  \caption{Segmentation performance comparison for the Cardiac MR $\rightarrow$ CT adaptation task. Starred methods perform source-free adaptation. Bolded cells show best performance.}
  \label{table:results-cardiac}
  
\end{table}

\begin{table}[ht!]
  \begin{adjustbox}{center}
    \scalebox{.77}{\small
    \setlength\tabcolsep{1.5pt} 
        \begin{tabular} { |c|cccc|c|cccc|c| }
            \hline
            &\multicolumn{5}{c|}{Dice} &\multicolumn{5}{c|}{Average Symmetric Surface Distance} \\ \hline
            Method &Liver &R.Kidney &L.Kidney &Spleen &Average &Liver &R.Kidney &L.Kidney &Spleen &Average \\ \hline
            Source-Only &73.1 &47.3 &57.3 &55.1 &58.2 &2.9 &5.6 &7.7 &7.4 &5.9  \\ \hline
            Supervised &94.2 &87.2 &88.9 &89.1 &89.8 &1.2 &1.2 &1.1 &1.7 &1.3 \\ \hline
            SynSeg-Net \cite{Huo_2019} &85.0 &82.1 &72.7 &81.0 &80.2 &2.2 &1.3 &2.1 &2.0 &1.9 \\ \hline
            AdaOutput \cite{Tsai_adaptseg_2018} &85.4 &79.7 &79.7 &81.7 &81.6 &1.7 &1.2 &1.8 &1.6 &1.6 \\ \hline
            CycleGAN \cite{zhu2020unpaired} &83.4 &79.3 &79.4 &77.3 &79.9 &1.8 &1.3 &\textbf{1.2} &1.9 &1.6 \\ \hline
            CyCADA \cite{hoffman2018cycada} &84.5 &78.6 &80.3 &76.9 &80.1 &2.6 &1.4 &1.3 &1.9 &1.8 \\ \hline
            SIFA \cite{chen2020unsupervised} &88.0 &\textbf{83.3} &\textbf{80.9} &\textbf{82.6} &\textbf{83.7} &\textbf{1.2} &\textbf{1.0} &1.5 &\textbf{1.6} &\textbf{1.3} \\ \hline
            \hline
            \textbf{SFS$^*$} &\textbf{88.3} &73.7 &80.7 &81.6 &81.1 &2.4 &4.1 &3.5 &2.7 &3.2 \\
            \hline
          \end{tabular}
    }
  \end{adjustbox}
  \caption{Segmentation performance comparison for the Abdominal MR $\rightarrow$ CT   task.}
  \label{table:results-abdomen}
  \vspace{-3mm}
\end{table}

\begin{figure*}[ht]
    \centering
    \includegraphics[width=.7\textwidth]{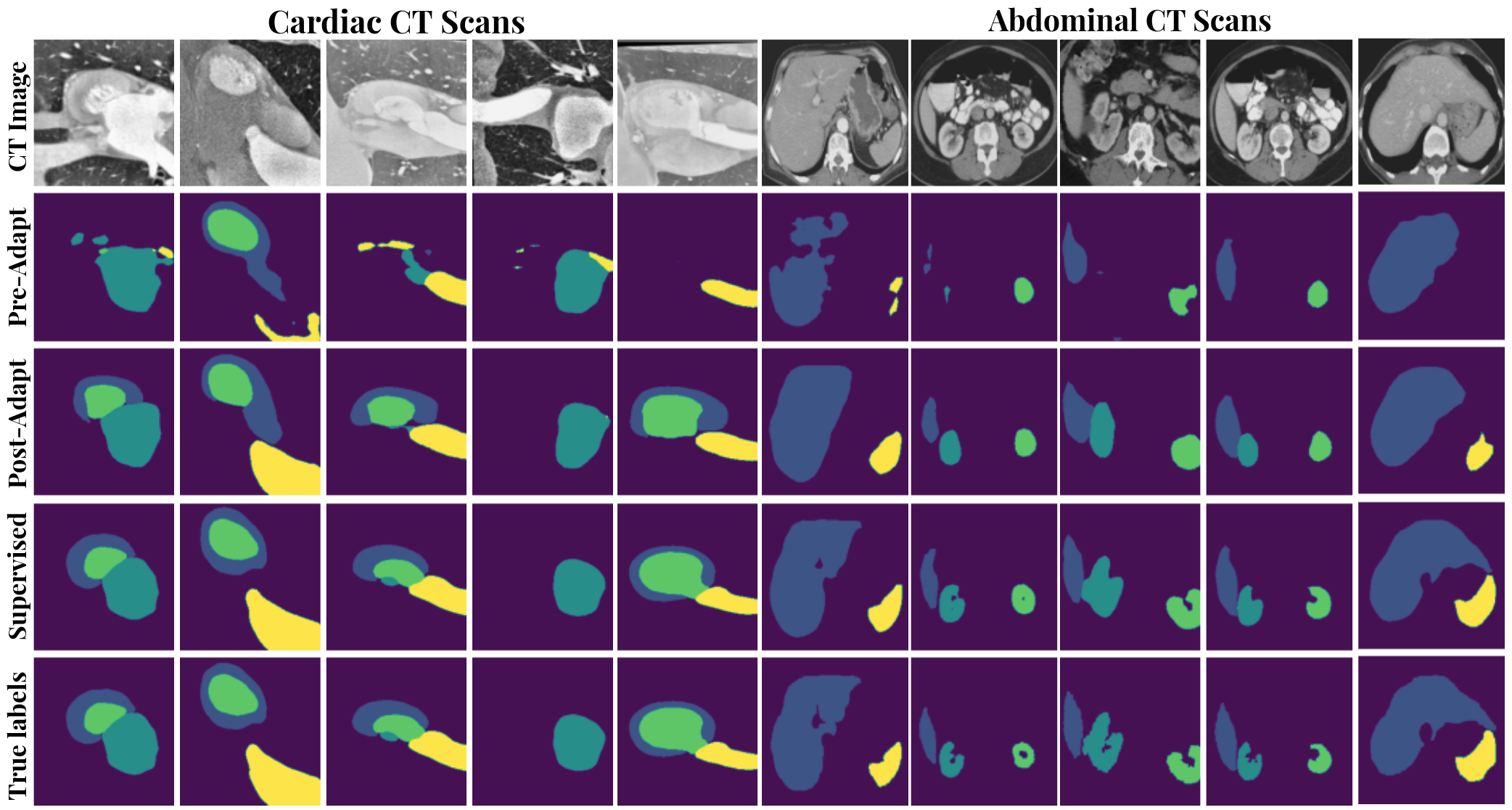}
    \caption{\small Segmentation maps of CT samples. The first five columns correspond to cardiac images, and last five correspond to abdominal images. From top to bottom: gray-scale CT images, source-only predictions, post-adaptation predictions, supervised predictions on the CT data, ground truth. }
    \label{figure:seg-maps}
      \vspace{-8mm}
\end{figure*}

In Figure~\ref{figure:seg-maps}, we present the improvement in segmentation on CT scans from  both datasets. In both cases, the supervised models are able to obtain a near perfect visual similarity to the ground truth segmentation which represent the upper-bound performance. Post-adaptation quality of the segmentation maps becomes much closer to the the supervised regime from a visual perspective. We observe fine details on image borders need more improvement in images $2,5,6,10$. This is in line with the observed ASSD performance. Overall, our approach offers significant gains with respect to the Dice coefficient, which directly measures the segmentation accuracy. The improvement in surface distance is also consistent, however best ASSD performance is observed for \cite{tomar2021self}, a method with joint access to source data. Still, our algorithm has the advantage of also maintaining data confidentiality during adaptation.
 
\subsection{Ablation Studies and Empirical Analysis}

\begin{figure}[!htb]
    \centering
    \subfigure[GMM samples]{
        \includegraphics[width=.2\textwidth]{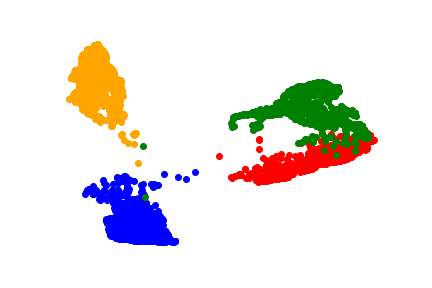}
        \label{figure:mmwhs-latent-features1}
    }
    \subfigure[Pre-adaptation]{
        \includegraphics[width=.2\textwidth]{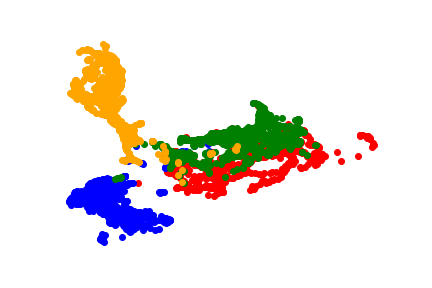}
        \label{figure:mmwhs-latent-features2}
    }
    \subfigure[Post-adaptation]{
        \includegraphics[width=.2\textwidth]{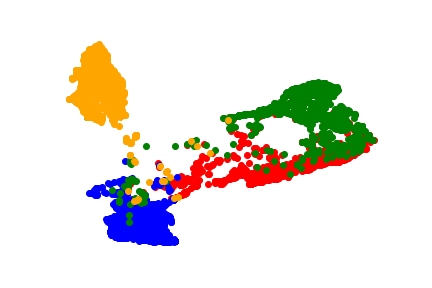}
        \label{figure:mmwhs-latent-features3}
    }
    \caption{\small Indirect distribution matching in the embedding space: (a) GMM samples approximating the MMWHS MR latent distribution, (b) CT latent embedding prior to adaptation (c) CT latent embedding post domain alignment. Colors correspond to: \textcolor{orange}{AA}, \textcolor{blue}{LAC}, \textcolor{green}{LVC}, \textcolor{red}{MYO}.}
    \label{figure:mmwhs-latent-features}
    \vspace{-3mm}
\end{figure}

\begin{figure}[!htb]
    \centering
    \subfigure[GMM samples]{
        \includegraphics[width=.2\textwidth]{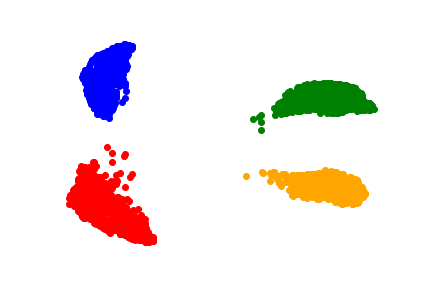}
        \label{figure:abdomen-latent-features1}
    }
    \subfigure[Pre-adaptation]{
        \includegraphics[width=.2\textwidth]{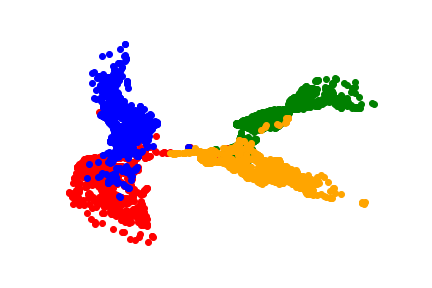}
        \label{figure:abdomen-latent-features2}
    }
    \subfigure[Post-adaptation]{
        \includegraphics[width=.2\textwidth]{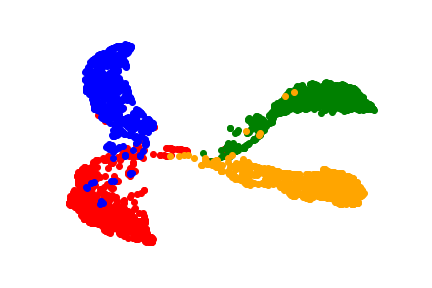}
        \label{figure:abdomen-latent-features3}
    }
    \caption{\small Indirect distribution matching: (a) GMM samples approximating the CHAOS MR latent distribution, (b) Multi-Atlas CT embedding prior to adaptation (c) Multi-Atlas CT embedding post adaptation. Colors correspond to: \textcolor{red}{liver}, \textcolor{blue}{right kidney}, \textcolor{green}{left kidney}, \textcolor{orange}{spleen}.} 
    \label{figure:abdomen-latent-features}
      \vspace{-3mm}
\end{figure}

We empirically demonstrate   why our algorithm works by screening changes in the latent embedding before and after adaptation. 
To visualize the embeddings, we use UMAP~\cite{mcinnes2020umap}    to reduce the high-dimensional  embeddings to $2D$. 
Figures~\ref{figure:mmwhs-latent-features} and ~\ref{figure:abdomen-latent-features} showcase the impact of our algorithm on the latent distribution of the two  datasets. In Figure~\ref{figure:mmwhs-latent-features1}, we record the latent embedding of  the GMM   distribution that is learned on the cardiac MR embeddings. Figure \ref{figure:mmwhs-latent-features2} exemplifies the distribution of the target CT samples before adaptation. We   see from Table \ref{table:results-cardiac} that the source-trained model is able to achieve some level of pre-adaptation class separation  which is confirmed in Figure \ref{figure:mmwhs-latent-features2}. 
In Figure \ref{figure:mmwhs-latent-features3} we observe that this overlap is reduced after adaptation. We also observe  that the latent embedding of the target CT samples is shifted towards the internal distribution, making the source-trained classifier generalizable. We repeat the same analysis for the organ segmentation dataset, and observe a similar outcome. 
We conclude that our intuition is confirmed, and the algorithm mitigates domain shift by performing distribution matching in the latent embedding space.

\begin{figure}[!htb]
    \centering
    \subfigure[$\rho=0$]{
        \includegraphics[width=.2\textwidth]{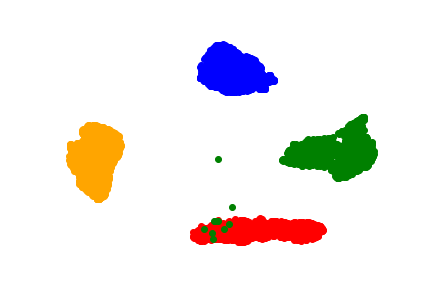}
        \label{figure:gaussians-0}
    }
    \subfigure[$\rho=0.8$]{
        \includegraphics[width=.2\textwidth]{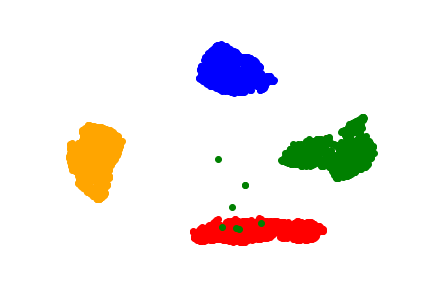}
        \label{figure:gaussians-8}
    }
    \subfigure[$\rho=.97$]{
        \includegraphics[width=.2\textwidth]{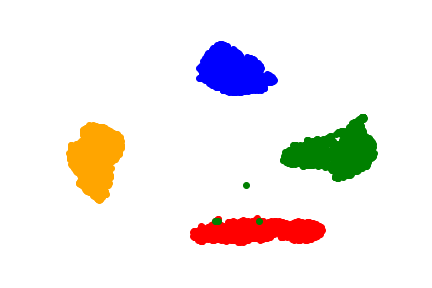}
        \label{figure:gaussians-97}
    }
    \caption{\small Learnt Gaussian embeddings on the cardiac dataset for different $\rho$.}
    \label{figure:mmwhs-gaussian-embeddings}
   \vspace{-3mm}
\end{figure}

We also investigate the impact of the $\rho$ parameter on our internal distribution. In Figure \ref{figure:mmwhs-gaussian-embeddings} we present the UMAP visualization  for the learnt GMM embeddings for three different values of $\rho$. We observe that while some classes will be separated for $\rho=0$,  using high confidence samples to learn the GMM will yield higher separability in the internal distribution. We observe our algorithm is robust when  $\rho$ is close to $1$, hence our choice of $\rho=.97$. 

\begin{table}[ht!]
  \begin{adjustbox}{center}
    \scalebox{.75}{\small
    \setlength\tabcolsep{1.5pt} 
        \begin{tabular} {|c| *{5}{|ccc|}}
            \hline
            &\multicolumn{3}{|c||}{Ignore} &\multicolumn{3}{|c||}{MYO}
            &\multicolumn{3}{|c||}{LAC} &\multicolumn{3}{|c||}{LVC}
            &\multicolumn{3}{|c|}{AA} \\ \hline
            Ignore &\textbf{97.3} &\textbf{99.3} &\textbf{99.3} &\textbf{1.5} &\textbf{20.3} &\textbf{70.0} &0.2 &80.2 &14.8 &0.9 &6.2 &76.1 &0.2 &43.8 &51.7 \\ \hline
MYO &\textbf{13.2} &\textbf{10.4} &\textbf{89.5} &\textbf{81.6} &\textbf{72.2} &\textbf{72.2} &0.1 &52.7 &0.4 &\textbf{5.2} &\textbf{44.6} &\textbf{54.1} &0.0 &0.0 &0.0 \\ \hline
LAC &\textbf{15.1} &\textbf{45.4} &\textbf{46.3} &\textbf{2.5} &\textbf{2.6} &\textbf{79.7} &\textbf{76.1} &\textbf{88.4} &\textbf{88.4} &\textbf{5.9} &\textbf{7.4} &\textbf{87.4} &0.4 &5.8 &77.0 \\ \hline
LVC &0.6 &67.7 &2.3 &\textbf{16.5} &\textbf{33.4} &\textbf{66.3} &0.2 &83.8 &13.0 &\textbf{82.7} &\textbf{92.4} &\textbf{92.4} &0.0 &93.3 &0.0 \\ \hline
AA &\textbf{18.5} &\textbf{7.8} &\textbf{90.9} &0.0 &0.0 &43.7 &\textbf{1.3} &\textbf{5.7} &\textbf{6.2} &0.1 &0.0 &12.9 &\textbf{80.1} &\textbf{91.2} &\textbf{91.2} \\ \hline
        \end{tabular}
    }
  \end{adjustbox}
  \caption{\small Percentage of shift in pixel labels during adaptation for the cardiac dataset. A cell $(i,j)$ in the table has three values. The first value represents the percentage of pixels labeled $i$ that are labeled $j$ after adaptation. The second value represents the percentage of switching pixels whose true label is $i$ - lower is better. The third value represents the percentage of switching pixels whose true label is $j$ - higher is better. Bolded cells denote label shift where more than $1\%$ of pixels migrate from $i$ to $j$.}
  \label{table:mmwhs-percentage-migrate}
\end{table}

\begin{table}[ht!]
  \begin{adjustbox}{center}
    \scalebox{.72}{\small
    \setlength\tabcolsep{1.5pt} 
        \begin{tabular} {|c| *{5}{|ccc|}}
            \hline
            &\multicolumn{3}{|c||}{Ignore} &\multicolumn{3}{|c||}{Liver}
            &\multicolumn{3}{|c||}{R. Kidney} &\multicolumn{3}{|c||}{L. Kidney}
            &\multicolumn{3}{|c|}{Spleen} \\ \hline
            Ignore &\textbf{94.6} &\textbf{98.4} &\textbf{98.4} &\textbf{3.0} &\textbf{18.0} &\textbf{81.6} &0.7 &23.5 &74.3 &0.7 &34.9 &62.6 &\textbf{1.0} &\textbf{19.3} &\textbf{80.5} \\ \hline
Liver &\textbf{6.6} &\textbf{38.1} &\textbf{60.8} &\textbf{92.6} &\textbf{91.3} &\textbf{91.3} &0.8 &10.4 &55.1 &0.0 &0.0 &0.0 &0.0 &39.0 &10.2 \\ \hline
R.Kidney &\textbf{5.0} &\textbf{13.1} &\textbf{86.9} &0.2 &0.0 &76.9 &\textbf{94.8} &\textbf{94.7} &\textbf{94.7} &0.0 &0.0 &0.0 &0.0 &0.0 &0.0 \\ \hline
L.Kidney &\textbf{2.2} &\textbf{24.2} &\textbf{75.0} &0.1 &0.0 &0.0 &0.0 &23.7 &0.0 &\textbf{97.5} &\textbf{87.8} &\textbf{87.8} &0.2 &0.0 &7.2 \\ \hline
Spleen &\textbf{23.1} &\textbf{20.8} &\textbf{79.2} &0.1 &20.2 &0.0 &0.2 &75.0 &0.0 &0.0 &69.4 &0.0 &\textbf{76.6} &\textbf{78.7} &\textbf{78.7} \\ \hline
        \end{tabular}
    }
  \end{adjustbox}
  \caption{\small Percentage of shift in pixel labels during adaptation for the abdominal organ dataset. The same methodology as in Table \ref{table:mmwhs-percentage-migrate} is used.}
  \label{table:abdomen-percentage-migrate}
  \vspace{-4mm}
\end{table}

The outcome of pixel label shift is analyzed in Tables \ref{table:mmwhs-percentage-migrate} and \ref{table:abdomen-percentage-migrate}. In Table \ref{table:mmwhs-percentage-migrate} we observe that for the cardiac dataset there exists significant inter-class label transfer, for approximately $20\%$ of pixels, evenly distributed across classes. We see the majority of these shifts leading to an improvement in labeling accuracy, including all shifts where at least $1\%$ of labels migrate, which is in line with our other reported results. These findings also corroborate with our observed embeddings. We can see from Table \ref{table:mmwhs-percentage-migrate} that during adaptation there is significant label migration between \textit{LVC} and \textit{MYO}, and this can be observed in the increased separation between the two classes in Figures \ref{figure:mmwhs-latent-features2} and  \ref{figure:mmwhs-latent-features3}. For the abdominal organ dataset we observe significantly less label shift between classes, with most of the activity involving previously labeled pixels being correctly le-labeled as \textit{Ignore} after adaptation, or pixels initially in \textit{Ignore} being correctly le-labeled to their appropriate class. 

We also perform an ablative experiment for the $\omega$ parameter using the the cardiac dataset in Table \ref{table:results-omega-components}. We observe a large increase in performance when using more than one component per class. However, this benefit decreases as more components are employed. We observe using  more than $2$ components increases the Dice score, and more than a $30\%$ drop in ASSD. We conclude a larger number of class components can offer a more expressive approximation of the source distribution, leading to improvements for segmentation accuracy and organ border quality. In our study we choose $\omega=3$ to balance performance and complexity. 

Full experimental setup and additional results are provided in the appendix.

\begin{table}[ht!]
  \begin{adjustbox}{center}
    \scalebox{.9}{
    \small
    \setlength\tabcolsep{1.5pt} 
        \begin{tabular} { |c|cccc|c|cccc|c| }
            \hline
            &\multicolumn{5}{c|}{Dice} &\multicolumn{5}{c|}{Average Symmetric Surface Distance} \\ \hline
            $\omega$-SFS &AA &LAC &LVC &MYO &Average &AA &LAC &LVC &MYO &Average \\ \hline
            \textbf{1-SFS} &86.2 &83.5 &75.4 &70.9 &79.0 &11.1 &5.0 &10.8 &3.6 &9.8 \\
            \hline
            \textbf{3-SFS} &88.0 &83.7 &81.0 &72.5 &81.3 &6.3 &7.2 &4.7 &6.1 &6.1 \\
            \hline
            \textbf{5-SFS} &88.0 &83.8 &81.9 &73.3 &81.7 &6.2 &7.4 &4.8 &5.7 &6.0 \\
            \hline
            \textbf{7-SFS} &86.8 &84.8 &82.0 &73.5 &81.8 &4.8 &7.2 &4.4 &5.6 &5.9 \\
            \hline
          \end{tabular}
    }
  \end{adjustbox}
  \caption{\small Segmentation performance versus $\omega$ for the Cardiac MR $\rightarrow$ CT adaptation task. 
  }
  \label{table:results-omega-components}
    \vspace{-3mm}
\end{table}

\section{Conclusion}
\label{sec:conclusion}

We developed a novel UDA algorithm for  semantic segmentation of confidential medical data. 
Our idea is based on estimating the source internal distribution via a GMM and then using is to align source and target domains indirectly. We provided a empirical analysis to demonstrate why our method is effective and it leads to competitive performance on two real-world datasets   when compared to state of the art approaches in medical semantic segmentation that require joint access to source and target data for adaptation.

\clearpage

\section{Appendix}
\label{sec:appendix}

\subsection{Experimental Setup}

We use the same network architecture on both the cardiac and organ image segmentation UDA task. We use a DeepLabV3 \cite{chen2017deeplab} feature extractor with a VGG16 backbone \cite{simonyan2015deep}, followed by a one layer classifier. 

\par We train the network on the supervised source samples with a training schedule of $30,000$ epochs repeated $3$ times. The optimizer of choice is Adam with learning rate $1e-4$, $\epsilon=1e-6$ and decay of $1e-6$. We use the standard pixel-wise cross entropy loss, and batch size of $16$. For the abdominal organ segmentation dataset, we observed better performance by using a weighted cross entropy loss.

\par We learn the empirical internal distribution using a parameter $\rho=.97$. We observed good separability in the latent distribution for $\rho \geq .9$. 

\par We use $\omega=3$ components per each of the $K$ classes, though as seen in Table \ref{table:results-omega-components}, a larger $\omega$ could potentially lead to further performance gains. $\omega=3$ strikes a balance between the complexity of the GMM model and realized performance.

\par Finally, when performing adaptation, we performed $35,000$ epochs of training, with a batch size of $32$. We again use an Adam optimizer with a learning rate of $5e-5$, $\epsilon=1e-1$ and decay of $1e-6$. Due to GPU memory constraints leading to a limited amount of image slices per batch, and therefore a large label distribution shifts between target batches, when sampling from the learnt GMMs we approximate the target distribution via the batch label distribution. 

\par Experiments were done on a NVIDIA RTX 3090 GPU. Code is provided in the supplementary material section of this submission, and will be made freely available online at a later date. 

\subsection{Additional Ablation Studies}

We further empirically analyze different components of our approach to demonstrate their effectiveness.

\par \textbf{Fine-tuning the classifier.} As we discussed in the main body of the paper, after learning an internal distribution characterizing the source embeddings, we align the target embeddings to this distribution by minimizing Sliced Wasserstein Distance. In addition, we also further train the classifier on samples from this distribution to account for differences to the original source embedding distribution. We next discuss the benefit of fine tuning the classifier, based on the results in Table \ref{table:results-classifier-fine-tuning}.

\begin{table}[htb!]
  \begin{adjustbox}{center}
    \scalebox{.8}{
        \begin{tabular} { |c|c|c| }
            \hline
            Metric &Fine-Tuned Classifier &Source Domain Classifier\\ \hline
            Dice &\textbf{81.3} &80.9\\ \hline
            ASSD &\textbf{6.1} &7.35\\ \hline
          \end{tabular}
    }
  \end{adjustbox}
  \caption{Target performance on the MMWHS adaptation task of our method with and without fine tuning the classifier on samples from the internal distribution. Bolded values indicate best performance.}
  \label{table:results-classifier-fine-tuning}
\end{table}

Given the learnt empirical means and covariances for the internal distribution, we compare the performance after target domain adaptation between a model that fine tunes the classifier and a model that does not update the classifier after source training. As expected, fine tuning the classifier offers a prediction boost, even if the difference is not a significant one. The internal distribution is meant to encourage the target embeddings to share a similar latent space with the source embeddings, and fine tuning the classifier accounts for the distribution shift between the source embeddings and learnt internal distribution.

\begin{figure*}[!htb]
    \centering
    \includegraphics[width=.9\textwidth]{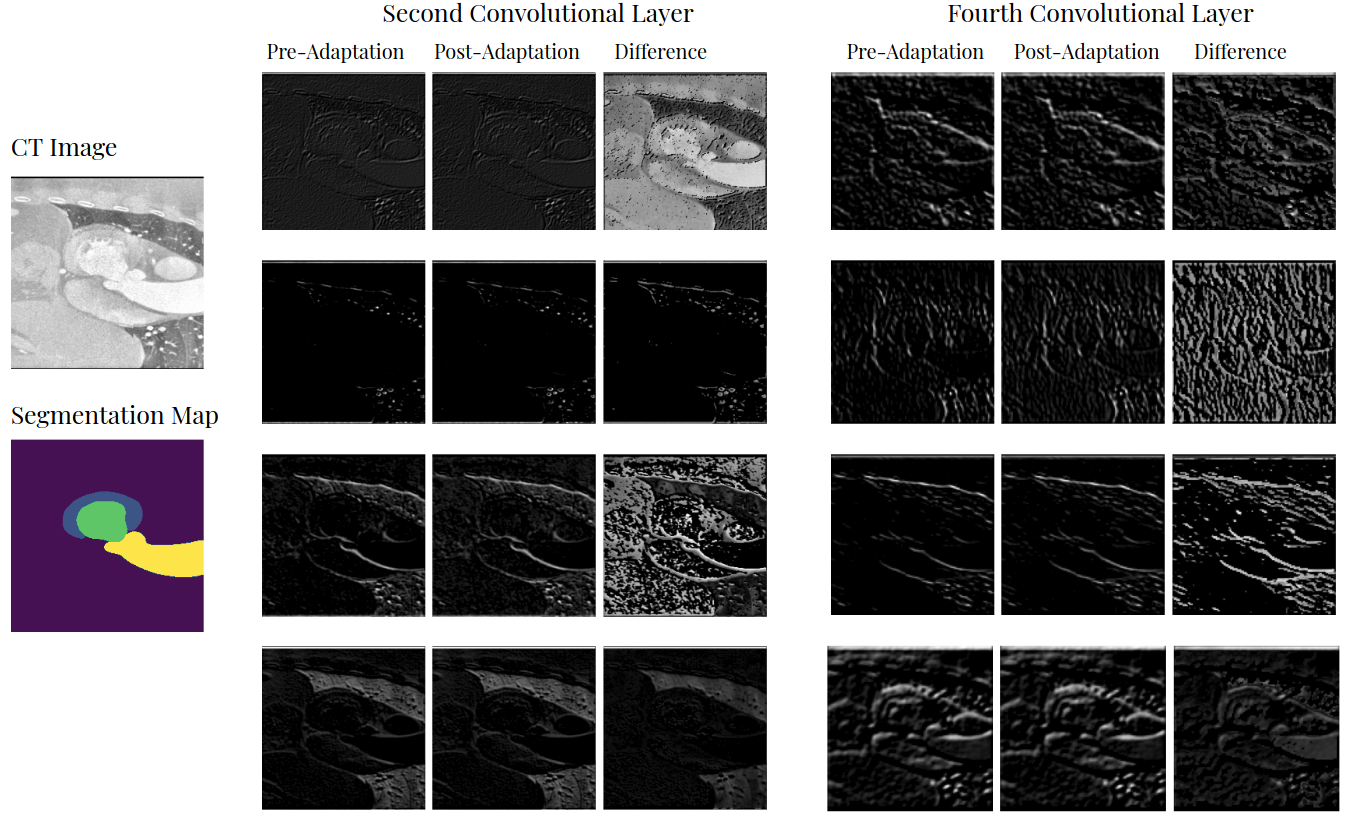}
    \caption{Filter maps of a cardiac CT image before and after model adaptation. In the case of filter differences in absolute value, dark grey symbolizes lower values, while light gray symbolizes higher values.}
    \label{figure:filters}
\end{figure*}

\textbf{Filter visualization}. We also investigate the information encoded in the convolutional filters before and after adaptation. Based on our results, we expect network filters to retain most of their structure from source training, and not alter this structure too much during distribution matching. We exemplify this in Figure \ref{figure:filters}. We record the visual characteristics of the network filters after the first two convolutional layers and the first four convolutional layers. We observe filters appear visually similar before and after adaptation, signifying image structural features learnt by the network do not undergo significant change, even though changes in filter values can be observed under the \textit{Difference} columns. 

\begin{thebibliography}{100}
\providecommand{\natexlab}[1]{#1}
\providecommand{\url}[1]{\texttt{#1}}
\expandafter\ifx\csname urlstyle\endcsname\relax
  \providecommand{\doi}[1]{doi: #1}\else
  \providecommand{\doi}{doi: \begingroup \urlstyle{rm}\Url}\fi

\bibitem[Ayache(2017)]{AYACHE2017xxiii}
Nicholas Ayache.
\newblock Deep learning for medical image analysis.
\newblock In S.~Kevin Zhou, Hayit Greenspan, and Dinggang Shen, editors,
  \emph{Deep Learning for Medical Image Analysis}, page xxiii. Academic Press,
  2017.
\newblock ISBN 978-0-12-810408-8.
\newblock \doi{https://doi.org/10.1016/B978-0-12-810408-8.00030-4}.
\newblock URL
  \url{http://www.sciencedirect.com/science/article/pii/B9780128104088000304}.

\bibitem[Bateson et~al.(2020)Bateson, Kervadec, Dolz, Lombaert, and
  Ben~Ayed]{BatesonSFDA}
Mathilde Bateson, Hoel Kervadec, Jose Dolz, Herv{\'e} Lombaert, and Ismail
  Ben~Ayed.
\newblock Source-relaxed domain adaptation for image segmentation.
\newblock In \emph{Medical Image Computing and Computer Assisted Intervention
  -- MICCAI 2020}, pages 490--499, Cham, 2020. Springer International
  Publishing.

\bibitem[Bateson et~al.(2022)Bateson, Kervadec, Dolz, Lombaert, and
  Ayed]{bateson2022source}
Mathilde Bateson, Hoel Kervadec, Jose Dolz, Herv{\'e} Lombaert, and Ismail~Ben
  Ayed.
\newblock Source-free domain adaptation for image segmentation.
\newblock \emph{Medical Image Analysis}, page 102617, 2022.

\bibitem[Bertinetto et~al.(2016)Bertinetto, Valmadre, Henriques, Vedaldi, and
  Torr]{bertinetto2016fully}
Luca Bertinetto, Jack Valmadre, Jo{\~a}o~F. Henriques, Andrea Vedaldi, and
  Philip H.~S. Torr.
\newblock Fully-convolutional siamese networks for object tracking.
\newblock In Gang Hua and Herv{\'e} J{\'e}gou, editors, \emph{Computer Vision
  -- ECCV 2016 Workshops}, pages 850--865, Cham, 2016. Springer International
  Publishing.
\newblock ISBN 978-3-319-48881-3.

\bibitem[Bousmalis et~al.(2017)Bousmalis, Silberman, Dohan, Erhan, and
  Krishnan]{bousmalis2017unsupervised}
Konstantinos Bousmalis, Nathan Silberman, David Dohan, Dumitru Erhan, and Dilip
  Krishnan.
\newblock Unsupervised pixel-level domain adaptation with generative
  adversarial networks.
\newblock In \emph{Proceedings of the IEEE conference on computer vision and
  pattern recognition}, pages 3722--3731, 2017.

\bibitem[{Chen} et~al.(2020){Chen}, {Dou}, {Chen}, {Qin}, and
  {Heng}]{chen2020unsupervised}
C.~{Chen}, Q.~{Dou}, H.~{Chen}, J.~{Qin}, and P.~A. {Heng}.
\newblock Unsupervised bidirectional cross-modality adaptation via deeply
  synergistic image and feature alignment for medical image segmentation.
\newblock \emph{IEEE Transactions on Medical Imaging}, 39\penalty0
  (7):\penalty0 2494--2505, 2020.
\newblock \doi{10.1109/TMI.2020.2972701}.

\bibitem[Chen et~al.(2019{\natexlab{a}})Chen, Xie, Huang, Rong, Ding, Huang,
  Xu, and Huang]{chen2019progressive}
Chaoqi Chen, Weiping Xie, Wenbing Huang, Yu~Rong, Xinghao Ding, Yue Huang,
  Tingyang Xu, and Junzhou Huang.
\newblock Progressive feature alignment for unsupervised domain adaptation.
\newblock In \emph{Proceedings of the IEEE Conference on Computer Vision and
  Pattern Recognition}, pages 627--636, 2019{\natexlab{a}}.

\bibitem[Chen et~al.(2018)Chen, Dou, Chen, and Heng]{chen2018semantic}
Cheng Chen, Qi~Dou, Hao Chen, and Pheng-Ann Heng.
\newblock Semantic-aware generative adversarial nets for unsupervised domain
  adaptation in chest x-ray segmentation.
\newblock In \emph{International workshop on machine learning in medical
  imaging}, pages 143--151. Springer, 2018.

\bibitem[Chen et~al.(2019{\natexlab{b}})Chen, Dou, Chen, Qin, and
  Heng]{chen2019synergistic}
Cheng Chen, Qi~Dou, Hao Chen, Jing Qin, and Pheng-Ann Heng.
\newblock Synergistic image and feature adaptation: Towards cross-modality
  domain adaptation for medical image segmentation.
\newblock In \emph{Proceedings of The Thirty-Third Conference on Artificial
  Intelligence (AAAI)}, pages 865--872, 2019{\natexlab{b}}.

\bibitem[Chen et~al.(2017)Chen, Papandreou, Kokkinos, Murphy, and
  Yuille]{chen2017deeplab}
Liang-Chieh Chen, George Papandreou, Iasonas Kokkinos, Kevin Murphy, and Alan~L
  Yuille.
\newblock Deeplab: Semantic image segmentation with deep convolutional nets,
  atrous convolution, and fully connected crfs.
\newblock \emph{IEEE transactions on pattern analysis and machine
  intelligence}, 40\penalty0 (4):\penalty0 834--848, 2017.

\bibitem[Chen et~al.(2020)Chen, Lian, Wang, Deng, Kuang, Fung, Gateno, Yap,
  Xia, and Shen]{chen2020anatomy}
Xu~Chen, Chunfeng Lian, Li~Wang, Hannah Deng, Tianshu Kuang, Steve Fung, Jaime
  Gateno, Pew-Thian Yap, James~J Xia, and Dinggang Shen.
\newblock Anatomy-regularized representation learning for cross-modality
  medical image segmentation.
\newblock \emph{IEEE Transactions on Medical Imaging}, 40\penalty0
  (1):\penalty0 274--285, 2020.

\bibitem[Choi et~al.(2019)Choi, Kim, and Kim]{choi2019self}
Jaehoon Choi, Taekyung Kim, and Changick Kim.
\newblock Self-ensembling with gan-based data augmentation for domain
  adaptation in semantic segmentation.
\newblock In \emph{Proceedings of the IEEE international conference on computer
  vision}, pages 6830--6840, 2019.

\bibitem[Courty et~al.(2016)Courty, Flamary, Tuia, and
  Rakotomamonjy]{courty2016optimal}
Nicolas Courty, R{\'e}mi Flamary, Devis Tuia, and Alain Rakotomamonjy.
\newblock Optimal transport for domain adaptation.
\newblock \emph{IEEE transactions on pattern analysis and machine
  intelligence}, 39\penalty0 (9):\penalty0 1853--1865, 2016.

\bibitem[Dou et~al.(2017)Dou, Yu, Chen, Jin, Yang, Qin, and Heng]{DOU201740}
Qi~Dou, Lequan Yu, Hao Chen, Yueming Jin, Xin Yang, Jing Qin, and Pheng-Ann
  Heng.
\newblock 3d deeply supervised network for automated segmentation of volumetric
  medical images.
\newblock \emph{Medical Image Analysis}, 41:\penalty0 40 -- 54, 2017.
\newblock ISSN 1361-8415.
\newblock \doi{https://doi.org/10.1016/j.media.2017.05.001}.
\newblock URL
  \url{http://www.sciencedirect.com/science/article/pii/S1361841517300725}.
\newblock Special Issue on the 2016 Conference on Medical Image Computing and
  Computer Assisted Intervention (Analog to MICCAI 2015).

\bibitem[Dou et~al.(2018)Dou, Ouyang, Chen, Chen, and
  Heng]{dou2018unsupervised}
Qi~Dou, Cheng Ouyang, Cheng Chen, Hao Chen, and Pheng-Ann Heng.
\newblock Unsupervised cross-modality domain adaptation of convnets for
  biomedical image segmentations with adversarial loss.
\newblock In \emph{Proceedings of the 27th International Joint Conference on
  Artificial Intelligence (IJCAI)}, pages 691--697, 2018.

\bibitem[Dou et~al.(2019)Dou, Ouyang, Chen, Chen, Glocker, Zhuang, and
  Heng]{dou2019pnp}
Qi~Dou, Cheng Ouyang, Cheng Chen, Hao Chen, Ben Glocker, Xiahai Zhuang, and
  Pheng-Ann Heng.
\newblock Pnp-adanet: Plug-and-play adversarial domain adaptation network at
  unpaired cross-modality cardiac segmentation.
\newblock \emph{IEEE Access}, 7:\penalty0 99065--99076, 2019.

\bibitem[{Drossos} et~al.(2019){Drossos}, {Magron}, and
  {Virtanen}]{drossos2019unsupervised}
K.~{Drossos}, P.~{Magron}, and T.~{Virtanen}.
\newblock Unsupervised adversarial domain adaptation based on the wasserstein
  distance for acoustic scene classification.
\newblock In \emph{2019 IEEE Workshop on Applications of Signal Processing to
  Audio and Acoustics (WASPAA)}, pages 259--263, 2019.
\newblock \doi{10.1109/WASPAA.2019.8937231}.

\bibitem[Ghifary et~al.(2016)Ghifary, Kleijn, Zhang, Balduzzi, and
  Li]{ghifary2016deep}
Muhammad Ghifary, W~Bastiaan Kleijn, Mengjie Zhang, David Balduzzi, and Wen Li.
\newblock Deep reconstruction-classification networks for unsupervised domain
  adaptation.
\newblock In \emph{European Conference on Computer Vision}, pages 597--613.
  Springer, 2016.

\bibitem[Han et~al.(2021)Han, Qi, Yu, Zhou, Zheng, Shi, and Gao]{han2021deep}
Xiaoting Han, Lei Qi, Qian Yu, Ziqi Zhou, Yefeng Zheng, Yinghuan Shi, and Yang
  Gao.
\newblock Deep symmetric adaptation network for cross-modality medical image
  segmentation.
\newblock \emph{IEEE transactions on medical imaging}, 41\penalty0
  (1):\penalty0 121--132, 2021.

\bibitem[Hecker et~al.(2018)Hecker, Dai, and Van~Gool]{Hecker_2018_ECCV}
Simon Hecker, Dengxin Dai, and Luc Van~Gool.
\newblock End-to-end learning of driving models with surround-view cameras and
  route planners.
\newblock In \emph{Proceedings of the European Conference on Computer Vision
  (ECCV)}, September 2018.

\bibitem[Hoffman et~al.(2018)Hoffman, Tzeng, Park, Zhu, Isola, Saenko, Efros,
  and Darrell]{hoffman2018cycada}
Judy Hoffman, Eric Tzeng, Taesung Park, Jun-Yan Zhu, Phillip Isola, Kate
  Saenko, Alexei Efros, and Trevor Darrell.
\newblock Cycada: Cycle-consistent adversarial domain adaptation.
\newblock In \emph{International conference on machine learning}, pages
  1989--1998. PMLR, 2018.

\bibitem[Huo et~al.(2018)Huo, Xu, Bao, Assad, Abramson, and
  Landman]{huo2018adversarial}
Yuankai Huo, Zhoubing Xu, Shunxing Bao, Albert Assad, Richard~G Abramson, and
  Bennett~A Landman.
\newblock Adversarial synthesis learning enables segmentation without target
  modality ground truth.
\newblock In \emph{2018 IEEE 15th international symposium on biomedical imaging
  (ISBI 2018)}, pages 1217--1220. IEEE, 2018.

\bibitem[Huo et~al.(2019)Huo, Xu, Moon, Bao, Assad, Moyo, Savona, Abramson, and
  Landman]{Huo_2019}
Yuankai Huo, Zhoubing Xu, Hyeonsoo Moon, Shunxing Bao, Albert Assad, Tamara~K.
  Moyo, Michael~R. Savona, Richard~G. Abramson, and Bennett~A. Landman.
\newblock Synseg-net: Synthetic segmentation without target modality ground
  truth.
\newblock \emph{IEEE Transactions on Medical Imaging}, 38\penalty0
  (4):\penalty0 1016–1025, Apr 2019.
\newblock ISSN 1558-254X.
\newblock \doi{10.1109/tmi.2018.2876633}.
\newblock URL \url{http://dx.doi.org/10.1109/TMI.2018.2876633}.

\bibitem[Kamnitsas et~al.(2017{\natexlab{a}})Kamnitsas, Baumgartner, Ledig,
  Newcombe, Simpson, Kane, Menon, Nori, Criminisi, Rueckert, and
  Glocker]{10.1007/978-3-319-59050-9_47}
Konstantinos Kamnitsas, Christian Baumgartner, Christian Ledig, Virginia
  Newcombe, Joanna Simpson, Andrew Kane, David Menon, Aditya Nori, Antonio
  Criminisi, Daniel Rueckert, and Ben Glocker.
\newblock Unsupervised domain adaptation in brain lesion segmentation with
  adversarial networks.
\newblock In Marc Niethammer, Martin Styner, Stephen Aylward, Hongtu Zhu, Ipek
  Oguz, Pew-Thian Yap, and Dinggang Shen, editors, \emph{Information Processing
  in Medical Imaging}, pages 597--609, Cham, 2017{\natexlab{a}}. Springer
  International Publishing.

\bibitem[Kamnitsas et~al.(2017{\natexlab{b}})Kamnitsas, Baumgartner, Ledig,
  Newcombe, Simpson, Kane, Menon, Nori, Criminisi, Rueckert,
  et~al.]{kamnitsas2017unsupervised}
Konstantinos Kamnitsas, Christian Baumgartner, Christian Ledig, Virginia
  Newcombe, Joanna Simpson, Andrew Kane, David Menon, Aditya Nori, Antonio
  Criminisi, Daniel Rueckert, et~al.
\newblock Unsupervised domain adaptation in brain lesion segmentation with
  adversarial networks.
\newblock In \emph{International conference on information processing in
  medical imaging}, pages 597--609. Springer, 2017{\natexlab{b}}.

\bibitem[Kavur et~al.(2019)Kavur, Selver, Dicle, Bar{\i}s, and
  Gezer]{CHAOSdata2019}
Ali~Emre Kavur, M~Alper Selver, Oguz Dicle, Mustafa Bar{\i}s, and N~Sinem
  Gezer.
\newblock Chaos-combined (ct-mr) healthy abdominal organ segmentation challenge
  data, Apr 2019.
\newblock URL \url{https://doi.org/10.5281/zenodo.3362844}.

\bibitem[Kazeminia et~al.(2020)Kazeminia, Baur, Kuijper, {van Ginneken}, Navab,
  Albarqouni, and Mukhopadhyay]{KAZEMINIA2020101938}
Salome Kazeminia, Christoph Baur, Arjan Kuijper, Bram {van Ginneken}, Nassir
  Navab, Shadi Albarqouni, and Anirban Mukhopadhyay.
\newblock Gans for medical image analysis.
\newblock \emph{Artificial Intelligence in Medicine}, 109:\penalty0 101938,
  2020.
\newblock ISSN 0933-3657.
\newblock \doi{https://doi.org/10.1016/j.artmed.2020.101938}.
\newblock URL
  \url{http://www.sciencedirect.com/science/article/pii/S0933365719311510}.

\bibitem[{Ker} et~al.(2018){Ker}, {Wang}, {Rao}, and {Lim}]{ker8241753deep}
J.~{Ker}, L.~{Wang}, J.~{Rao}, and T.~{Lim}.
\newblock Deep learning applications in medical image analysis.
\newblock \emph{IEEE Access}, 6:\penalty0 9375--9389, 2018.
\newblock \doi{10.1109/ACCESS.2017.2788044}.

\bibitem[Kim and Canny(2017)]{Kim_2017_ICCV}
Jinkyu Kim and John Canny.
\newblock Interpretable learning for self-driving cars by visualizing causal
  attention.
\newblock In \emph{Proceedings of the IEEE International Conference on Computer
  Vision (ICCV)}, Oct 2017.

\bibitem[Kundu et~al.(2020)Kundu, Venkat, Babu, et~al.]{kundu2020universal}
Jogendra~Nath Kundu, Naveen Venkat, R~Venkatesh Babu, et~al.
\newblock Universal source-free domain adaptation.
\newblock In \emph{Proceedings of the IEEE/CVF Conference on Computer Vision
  and Pattern Recognition}, pages 4544--4553, 2020.

\bibitem[Kundu et~al.(2021)Kundu, Kulkarni, Singh, Jampani, and
  Babu]{Kundu_2021_ICCV}
Jogendra~Nath Kundu, Akshay Kulkarni, Amit Singh, Varun Jampani, and
  R.~Venkatesh Babu.
\newblock Generalize then adapt: Source-free domain adaptive semantic
  segmentation.
\newblock In \emph{Proceedings of the IEEE/CVF International Conference on
  Computer Vision (ICCV)}, pages 7046--7056, October 2021.

\bibitem[Landman et~al.(2015)Landman, Xu, Igelsias, Styner, Langerak, and
  Klein]{landman2015multiatlas}
Bennett Landman, Z~Xu, JE~Igelsias, M~Styner, TR~Langerak, and A~Klein.
\newblock Multi-atlas labeling beyond the cranial vault-workshop and challenge,
  2015.

\bibitem[Le et~al.(2019)Le, Habrard, and Sebban]{LE2019249}
Tien-Nam Le, Amaury Habrard, and Marc Sebban.
\newblock Deep multi-wasserstein unsupervised domain adaptation.
\newblock \emph{Pattern Recognition Letters}, 125:\penalty0 249 -- 255, 2019.
\newblock ISSN 0167-8655.
\newblock \doi{https://doi.org/10.1016/j.patrec.2019.04.025}.
\newblock URL
  \url{http://www.sciencedirect.com/science/article/pii/S0167865519301400}.

\bibitem[LeCun et~al.(2015)LeCun, Bengio, and Hinton]{lecun2015deep}
Yann LeCun, Yoshua Bengio, and Geoffrey Hinton.
\newblock Deep learning.
\newblock \emph{nature}, 521\penalty0 (7553):\penalty0 436--444, 2015.

\bibitem[Lee et~al.(2019)Lee, Batra, Baig, and Ulbricht]{lee2019sliced}
Chen-Yu Lee, Tanmay Batra, Mohammad~Haris Baig, and Daniel Ulbricht.
\newblock Sliced wasserstein discrepancy for unsupervised domain adaptation.
\newblock In \emph{Proceedings of the IEEE Conference on Computer Vision and
  Pattern Recognition}, pages 10285--10295, 2019.

\bibitem[Lin et~al.(2017)Lin, Milan, Shen, and Reid]{lin2017refinenet}
Guosheng Lin, Anton Milan, Chunhua Shen, and Ian Reid.
\newblock Refinenet: Multi-path refinement networks for high-resolution
  semantic segmentation.
\newblock In \emph{Proceedings of the IEEE conference on computer vision and
  pattern recognition}, pages 1925--1934, 2017.

\bibitem[{Liu} et~al.(2020){Liu}, {Zhang}, {Song}, {Zhang}, {O’Donnell},
  {Huang}, {Chen}, and {Cai}]{liu2020pdam}
D.~{Liu}, D.~{Zhang}, Y.~{Song}, F.~{Zhang}, L.~{O’Donnell}, H.~{Huang},
  M.~{Chen}, and W.~{Cai}.
\newblock Pdam: A panoptic-level feature alignment framework for unsupervised
  domain adaptive instance segmentation in microscopy images.
\newblock \emph{IEEE Transactions on Medical Imaging}, pages 1--1, 2020.
\newblock \doi{10.1109/TMI.2020.3023466}.

\bibitem[Liu et~al.(2011)Liu, Xiong, Pulli, and Shapiro]{estimating2011liu}
Dingding Liu, Yingen Xiong, Kari Pulli, and Linda Shapiro.
\newblock Estimating image segmentation difficulty.
\newblock In \emph{International Workshop on Machine Learning and Data Mining
  in Pattern Recognition}, pages 484--495. Springer, 2011.

\bibitem[Long et~al.(2015)Long, Shelhamer, and Darrell]{long2015fully}
Jonathan Long, Evan Shelhamer, and Trevor Darrell.
\newblock Fully convolutional networks for semantic segmentation.
\newblock In \emph{Proceedings of the IEEE conference on computer vision and
  pattern recognition}, pages 3431--3440, 2015.

\bibitem[Ma et~al.(2019)Ma, Zhang, and Xu]{Ma_2019_CVPR}
Xinhong Ma, Tianzhu Zhang, and Changsheng Xu.
\newblock Gcan: Graph convolutional adversarial network for unsupervised domain
  adaptation.
\newblock In \emph{Proceedings of the IEEE/CVF Conference on Computer Vision
  and Pattern Recognition (CVPR)}, June 2019.

\bibitem[McInnes et~al.(2020)McInnes, Healy, and Melville]{mcinnes2020umap}
Leland McInnes, John Healy, and James Melville.
\newblock Umap: Uniform manifold approximation and projection for dimension
  reduction, 2020.

\bibitem[Motiian et~al.(2017)Motiian, Jones, Iranmanesh, and
  Doretto]{motiian2017few}
Saeid Motiian, Quinn Jones, Seyed Iranmanesh, and Gianfranco Doretto.
\newblock Few-shot adversarial domain adaptation.
\newblock In \emph{Advances in Neural Information Processing Systems}, pages
  6670--6680, 2017.

\bibitem[Noh et~al.(2015)Noh, Hong, and Han]{noh2015learning}
Hyeonwoo Noh, Seunghoon Hong, and Bohyung Han.
\newblock Learning deconvolution network for semantic segmentation.
\newblock In \emph{Proceedings of the IEEE international conference on computer
  vision}, pages 1520--1528, 2015.

\bibitem[Pan et~al.(2019)Pan, Yao, Li, Wang, Ngo, and
  Mei]{pan2019transferrable}
Yingwei Pan, Ting Yao, Yehao Li, Yu~Wang, Chong-Wah Ngo, and Tao Mei.
\newblock Transferrable prototypical networks for unsupervised domain
  adaptation.
\newblock In \emph{Proceedings of the IEEE Conference on Computer Vision and
  Pattern Recognition}, pages 2239--2247, 2019.

\bibitem[Qiu et~al.(2021)Qiu, Zhang, Lin, Niu, Liu, Du, and
  Tan]{qiu2021sourcefree}
Zhen Qiu, Yifan Zhang, Hongbin Lin, Shuaicheng Niu, Yanxia Liu, Qing Du, and
  Mingkui Tan.
\newblock Source-free domain adaptation via avatar prototype generation and
  adaptation, 2021.

\bibitem[Rostami(2021{\natexlab{a}})]{rostami2021lifelong}
Mohammad Rostami.
\newblock Lifelong domain adaptation via consolidated internal distribution.
\newblock \emph{Advances in Neural Information Processing Systems},
  34:\penalty0 11172--11183, 2021{\natexlab{a}}.

\bibitem[Rostami(2021{\natexlab{b}})]{rostami2021transfer}
Mohammad Rostami.
\newblock \emph{Transfer Learning Through Embedding Spaces}.
\newblock CRC Press, 2021{\natexlab{b}}.

\bibitem[Rostami(2022)]{rost2021unsupervised}
Mohammad Rostami.
\newblock Increasing model generalizability for unsupervised domain adaptation.
\newblock In \emph{Proceedings of the Conference on Lifelong Learning Agents},
  2022.

\bibitem[Rostami and Galstyan(2021)]{rostami2021domain}
Mohammad Rostami and Aram Galstyan.
\newblock Domain adaptation for sentiment analysis using increased intraclass
  separation.
\newblock \emph{arXiv preprint arXiv:2107.01598}, 2021.

\bibitem[Rostami et~al.(2018)Rostami, Huber, and Lu]{rostami2018crowdsourcing}
Mohammad Rostami, David Huber, and Tsai-Ching Lu.
\newblock A crowdsourcing triage algorithm for geopolitical event forecasting.
\newblock In \emph{Proceedings of the 12th ACM Conference on Recommender
  Systems}, pages 377--381, 2018.

\bibitem[Rostami et~al.(2019)Rostami, Kolouri, Eaton, and Kim]{rostami2019deep}
Mohammad Rostami, Soheil Kolouri, Eric Eaton, and Kyungnam Kim.
\newblock Deep transfer learning for few-shot sar image classification.
\newblock \emph{Remote Sensing}, 11\penalty0 (11):\penalty0 1374, 2019.

\bibitem[Saito et~al.(2018)Saito, Watanabe, Ushiku, and
  Harada]{saito2018maximum}
Kuniaki Saito, Kohei Watanabe, Yoshitaka Ushiku, and Tatsuya Harada.
\newblock Maximum classifier discrepancy for unsupervised domain adaptation.
\newblock In \emph{Proceedings of the IEEE Conference on Computer Vision and
  Pattern Recognition}, pages 3723--3732, 2018.

\bibitem[Saltori et~al.(2020)Saltori, Lathuili{\'e}re, Sebe, Ricci, and
  Galasso]{saltori2020sf}
Cristiano Saltori, St{\'e}phane Lathuili{\'e}re, Nicu Sebe, Elisa Ricci, and
  Fabio Galasso.
\newblock Sf-uda 3d: Source-free unsupervised domain adaptation for lidar-based
  3d object detection.
\newblock In \emph{2020 International Conference on 3D Vision (3DV)}, pages
  771--780. IEEE, 2020.

\bibitem[Sankaranarayanan et~al.(2018)Sankaranarayanan, Balaji, Castillo, and
  Chellappa]{sankaranarayanan2018generate}
Swami Sankaranarayanan, Yogesh Balaji, Carlos~D Castillo, and Rama Chellappa.
\newblock Generate to adapt: Aligning domains using generative adversarial
  networks.
\newblock In \emph{Proceedings of the IEEE Conference on Computer Vision and
  Pattern Recognition}, pages 8503--8512, 2018.

\bibitem[Shen et~al.(2017)Shen, Wu, and Suk]{shen2017deep}
Dinggang Shen, Guorong Wu, and Heung-Il Suk.
\newblock Deep learning in medical image analysis.
\newblock \emph{Annual Review of Biomedical Engineering}, 19\penalty0
  (1):\penalty0 221--248, 2017.
\newblock \doi{10.1146/annurev-bioeng-071516-044442}.
\newblock URL \url{https://doi.org/10.1146/annurev-bioeng-071516-044442}.
\newblock PMID: 28301734.

\bibitem[Simonyan and Zisserman(2015)]{simonyan2015deep}
Karen Simonyan and Andrew Zisserman.
\newblock Very deep convolutional networks for large-scale image recognition,
  2015.

\bibitem[Stan and Rostami(2021)]{stan2021unsupervised}
Serban Stan and Mohammad Rostami.
\newblock Unsupervised model adaptation for continual semantic segmentation.
\newblock In \emph{Proceedings of the AAAI Conference on Artificial
  Intelligence}, volume~35, pages 2593--2601, 2021.

\bibitem[Sun et~al.(2017{\natexlab{a}})Sun, Feng, and
  Saenko]{sun2017correlation}
Baochen Sun, Jiashi Feng, and Kate Saenko.
\newblock Correlation alignment for unsupervised domain adaptation.
\newblock In \emph{Domain Adaptation in Computer Vision Applications}, pages
  153--171. Springer, 2017{\natexlab{a}}.

\bibitem[Sun et~al.(2017{\natexlab{b}})Sun, Guo, Zhang, Li, Chen, Ma, Jin, Liu,
  Li, and Qian]{SUN201758}
Changjian Sun, Shuxu Guo, Huimao Zhang, Jing Li, Meimei Chen, Shuzhi Ma, Lanyi
  Jin, Xiaoming Liu, Xueyan Li, and Xiaohua Qian.
\newblock Automatic segmentation of liver tumors from multiphase
  contrast-enhanced ct images based on fcns.
\newblock \emph{Artificial Intelligence in Medicine}, 83:\penalty0 58 -- 66,
  2017{\natexlab{b}}.
\newblock ISSN 0933-3657.
\newblock \doi{https://doi.org/10.1016/j.artmed.2017.03.008}.
\newblock URL
  \url{http://www.sciencedirect.com/science/article/pii/S0933365716305930}.
\newblock Machine Learning and Graph Analytics in Computational Biomedicine.

\bibitem[Toldo et~al.(2020)Toldo, Maracani, Michieli, and
  Zanuttigh]{Toldo_Maracani_Michieli_Zanuttigh_2020}
Marco Toldo, Andrea Maracani, Umberto Michieli, and Pietro Zanuttigh.
\newblock Unsupervised domain adaptation in semantic segmentation: A review.
\newblock \emph{Technologies}, 8\penalty0 (2):\penalty0 35, Jun 2020.
\newblock ISSN 2227-7080.
\newblock \doi{10.3390/technologies8020035}.

\bibitem[Tomar et~al.(2021)Tomar, Lortkipanidze, Vray, Bozorgtabar, and
  Thiran]{tomar2021self}
Devavrat Tomar, Manana Lortkipanidze, Guillaume Vray, Behzad Bozorgtabar, and
  Jean-Philippe Thiran.
\newblock Self-attentive spatial adaptive normalization for cross-modality
  domain adaptation.
\newblock \emph{IEEE Transactions on Medical Imaging}, 2021.

\bibitem[Tsai et~al.(2018)Tsai, Hung, Schulter, Sohn, Yang, and
  Chandraker]{Tsai_adaptseg_2018}
Y.-H. Tsai, W.-C. Hung, S.~Schulter, K.~Sohn, M.-H. Yang, and M.~Chandraker.
\newblock Learning to adapt structured output space for semantic segmentation.
\newblock In \emph{IEEE Conference on Computer Vision and Pattern Recognition
  (CVPR)}, 2018.

\bibitem[Tzeng et~al.(2017)Tzeng, Hoffman, Saenko, and
  Darrell]{tzeng2017adversarial}
Eric Tzeng, Judy Hoffman, Kate Saenko, and Trevor Darrell.
\newblock Adversarial discriminative domain adaptation.
\newblock In \emph{Proceedings of the IEEE conference on computer vision and
  pattern recognition}, pages 7167--7176, 2017.

\bibitem[Venkateswara et~al.(2017)Venkateswara, Eusebio, Chakraborty, and
  Panchanathan]{venkateswara2017deep}
Hemanth Venkateswara, Jose Eusebio, Shayok Chakraborty, and Sethuraman
  Panchanathan.
\newblock Deep hashing network for unsupervised domain adaptation.
\newblock In \emph{Proceedings of the IEEE Conference on Computer Vision and
  Pattern Recognition}, pages 5018--5027, 2017.

\bibitem[Wu et~al.(2018)Wu, Han, Lin, Uzunbas, Goldstein, Lim, and
  Davis]{wu2018_dcan_eccv}
Zuxuan Wu, Xintong Han, Yen-Liang Lin, Mustafa~G{\"o}khan Uzunbas, Tom
  Goldstein, Ser~Nam Lim, and Larry~S. Davis.
\newblock Dcan: Dual channel-wise alignment networks for unsupervised scene
  adaptation.
\newblock In Vittorio Ferrari, Martial Hebert, Cristian Sminchisescu, and Yair
  Weiss, editors, \emph{Computer Vision -- ECCV 2018}, pages 535--552, Cham,
  2018. Springer International Publishing.

\bibitem[Yang et~al.(2021)Yang, Wang, van~de Weijer, Herranz, and
  Jui]{yang2021generalized}
Shiqi Yang, Yaxing Wang, Joost van~de Weijer, Luis Herranz, and Shangling Jui.
\newblock Generalized source-free domain adaptation, 2021.

\bibitem[Yilmaz et~al.(2006)Yilmaz, Javed, and Shah]{yilmaz2006survey}
Alper Yilmaz, Omar Javed, and Mubarak Shah.
\newblock Object tracking: A survey.
\newblock \emph{ACM Comput. Surv.}, 38\penalty0 (4):\penalty0 13–es, December
  2006.
\newblock ISSN 0360-0300.
\newblock \doi{10.1145/1177352.1177355}.
\newblock URL \url{https://doi.org/10.1145/1177352.1177355}.

\bibitem[Zhang et~al.(2019)Zhang, Chen, Huang, Lin, and Zhang]{zhang2019few}
Junyi Zhang, Ziliang Chen, Junying Huang, Liang Lin, and Dongyu Zhang.
\newblock Few-shot structured domain adaptation for virtual-to-real scene
  parsing.
\newblock In \emph{Proceedings of the IEEE International Conference on Computer
  Vision Workshops}, pages 0--0, 2019.

\bibitem[Zhang et~al.(2018{\natexlab{a}})Zhang, Ouyang, Li, and
  Xu]{Zhang_2018_CVPR}
Weichen Zhang, Wanli Ouyang, Wen Li, and Dong Xu.
\newblock Collaborative and adversarial network for unsupervised domain
  adaptation.
\newblock In \emph{Proceedings of the IEEE Conference on Computer Vision and
  Pattern Recognition (CVPR)}, June 2018{\natexlab{a}}.

\bibitem[Zhang et~al.(2018{\natexlab{b}})Zhang, Ouyang, Li, and
  Xu]{zhang2018collaborative}
Weichen Zhang, Wanli Ouyang, Wen Li, and Dong Xu.
\newblock Collaborative and adversarial network for unsupervised domain
  adaptation.
\newblock In \emph{Proceedings of the IEEE Conference on Computer Vision and
  Pattern Recognition}, pages 3801--3809, 2018{\natexlab{b}}.

\bibitem[Zhang et~al.(2017)Zhang, David, and Gong]{zhang2017_curriculum}
Yang Zhang, Philip David, and Boqing Gong.
\newblock Curriculum domain adaptation for semantic segmentation of urban
  scenes.
\newblock \emph{2017 IEEE International Conference on Computer Vision (ICCV)},
  Oct 2017.
\newblock \doi{10.1109/iccv.2017.223}.
\newblock URL \url{http://dx.doi.org/10.1109/ICCV.2017.223}.

\bibitem[Zhang et~al.(2018{\natexlab{c}})Zhang, Miao, Mansi, and
  Liao]{zhang2018task}
Yue Zhang, Shun Miao, Tommaso Mansi, and Rui Liao.
\newblock Task driven generative modeling for unsupervised domain adaptation:
  Application to x-ray image segmentation.
\newblock In \emph{International Conference on Medical Image Computing and
  Computer-Assisted Intervention}, pages 599--607. Springer,
  2018{\natexlab{c}}.

\bibitem[Zhao et~al.(2019)Zhao, Li, Wan, Sekuboyina, Hu, Tetteh, Piraud, and
  Menze]{zhao2019knowledge}
Yu~Zhao, Hongwei Li, Shaohua Wan, Anjany Sekuboyina, Xiaobin Hu, Giles Tetteh,
  Marie Piraud, and Bjoern Menze.
\newblock Knowledge-aided convolutional neural network for small organ
  segmentation.
\newblock \emph{IEEE journal of biomedical and health informatics}, 23\penalty0
  (4):\penalty0 1363--1373, 2019.

\bibitem[Zhou et~al.(2019)Zhou, Li, Bai, Wang, Chen, Han, Fishman, and
  Yuille]{zhou2019prioraware}
Yuyin Zhou, Zhe Li, Song Bai, Chong Wang, Xinlei Chen, Mei Han, Elliot Fishman,
  and Alan Yuille.
\newblock Prior-aware neural network for partially-supervised multi-organ
  segmentation, 2019.

\bibitem[Zhu et~al.(2018)Zhu, Yang, Liu, Kim, Zhang, and Yang]{Zhu_2018_ECCV}
Ji~Zhu, Hua Yang, Nian Liu, Minyoung Kim, Wenjun Zhang, and Ming-Hsuan Yang.
\newblock Online multi-object tracking with dual matching attention networks.
\newblock In \emph{Proceedings of the European Conference on Computer Vision
  (ECCV)}, September 2018.

\bibitem[Zhu et~al.(2020)Zhu, Park, Isola, and Efros]{zhu2020unpaired}
Jun-Yan Zhu, Taesung Park, Phillip Isola, and Alexei~A. Efros.
\newblock Unpaired image-to-image translation using cycle-consistent
  adversarial networks, 2020.

\bibitem[Zhuang and Shen(2016)]{ZHUANG201677}
Xiahai Zhuang and Juan Shen.
\newblock Multi-scale patch and multi-modality atlases for whole heart
  segmentation of mri.
\newblock \emph{Medical Image Analysis}, 31:\penalty0 77 -- 87, 2016.
\newblock ISSN 1361-8415.
\newblock \doi{https://doi.org/10.1016/j.media.2016.02.006}.
\newblock URL
  \url{http://www.sciencedirect.com/science/article/pii/S1361841516000219}.

\bibitem[Zou et~al.(2020)Zou, Zhu, and Yan]{ijcai2020-455}
Danbing Zou, Qikui Zhu, and Pingkun Yan.
\newblock Unsupervised domain adaptation with dual scheme fusion network for
  medical image segmentation.
\newblock In \emph{Proceedings of the Twenty-Ninth International Joint
  Conference on Artificial Intelligence, IJCAI-20, International Joint
  Conferences on Artificial Intelligence Organization}, pages 3291--3298, 2020.

\end{thebibliography}
\end{document}